# HRNET: AI-on-Edge for mask detection and social distancing calculation

Kinshuk Sengupta
Microsoft Corporation, India
*kisengup@microsoft.com*

Praveen Ranjan Srivastava
Indian Institute of Management Rohtak
*praveen.ranjan@iimrohtak.ac.in*

## Abstract

**Purpose**

The purpose of the paper is to provide innovative emerging technology framework for community to combat epidemic situations. The paper proposes a unique outbreak response system framework based on artificial intelligence and edge computing for citizen centric services to help track and trace people eluding safety policies like mask detection and social distancing measure in public or workplace setup. The framework further provides implementation guideline in industrial setup as well for governance and contact tracing tasks. The adoption will thus lead in smart city planning and development focusing on citizen health systems contributing to improved quality of life.

**Design/methodology/approach**

The conceptual framework presented is validated through quantitative data analysis via secondary data collection from researcher's public websites, GitHub repositories and renowned journals and further benchmarking were conducted for experimental results in Microsoft Azure cloud environment.

**Findings**

The study includes selective AI-models for benchmark analysis and were assessed on performance and accuracy in edge computing environment for large scale societal setup. Overall YOLO model Outperforms in object detection task and is faster enough for mask detection and HRNetV2 outperform semantic segmentation problem applied to solve social distancing task in AI-Edge inferencing environmental setup.

**Originality/Value**

The paper proposes new Edge-AI algorithm for building technology-oriented solutions for detecting mask in human movement and social distance. The paper enriches the technological advancement in artificial intelligence and edge-computing applied to problems in society and healthcare systems. The framework further equips government agency, system providers to design and constructs technology-oriented models in community setup to Increase the quality of life using emerging technologies into smart urban environments.

## 1. Introduction

The recent outbreak of coronavirus SARS-CoV-2 infection early detected in December 2019 in Wuhan, China (Wang and Yu, 2020). The magnitude of infectious spread has affected more than 3.2million peoples causing 239K deaths, according to the European Centre for Disease Prevention and Control. In the latest report, the total death caused by fever is 64.7% and 52.9% due to cough (Wang et al.). The role of flattering the curve via quarantine and preventing community spread by using a respiratory surgical mask or N95 mask have found significance in controlling the spread in previously published literature(Wang and Yu, 2020). The literature had shown evidence indicating the use of a surgical mask reduces the transmissibility per individual by preventing the droplets transmission in both laboratory and clinical contexts (Howarda et al., 2020). In the compliant scenario for industrial workplaces, airports, and places for community gathering possess the highest risk of spread without prevention. Public health authorities have approached to contain the virus spread via isolation, personal protection, and hygiene compliance(Semple and Cherrie, 2020), social distancing, contact tracing, and surveillance application (Bedford et al., 2020).

Recent surveys and literature have studied how to handle community gatherings to prevent the global spread of COVID-19. Several research gaps that need to address for the response to COVID-19 need to be discussed in the current situation, such as developing an ethical framework to control and contain spread in such states. This also involves devising appropriate ways to prevent and control the infection by identifying optimal personal protective equipment(PPE's), and after that, understand behavior among various vulnerable groups (J Bedford, D Enria, J Giesecke, DL Heymann, C Ihekweazu, G Kobinger, et al., 2020). Many recent research works propose policymakers to make masks as an official guideline to stop spread in the community (Howarda et al., 2020). Hence, the need to find sustainably designed AI-powered technology solution like robotics, self-explainable digital solutions(Srivastava, P.R., Sengupta, K., Kumar, A., Biswas, B. and Ishizaka, A., 2021) to tackle the post-pandemic situation in society and industrial setup needs drastic attention to counter the ripple effect of COVID-19 in economic circumstances. Policymakers and industries would need an efficient solution within the industrial and societal structure to trace and track people and prevent droplet spread(Greenhalgh T, Schmid MB, Czypionka T, Bassler D, Gruer L, 2020). An older research review conducted randomized studies-controlled trials using masks and found that a low-cost intervention would be useful to break the transmission of the respiratory viruses Jefferson T, et al. (2011). The summary of existing studies is depicted in below table 1.

**Table 1.** Compare and contract review implication and discovering significant scope for GAP analysis.

| S. No | Research Article | Research Objective | Methodology/Technique | Limitations | Implications |
|---|---|---|---|---|---|
| 1. | Defending against the Novel Coronavirus (COVID-19) Outbreak: How Can the Internet of Things (IoT) help to save the World? **DOI**:https://doi.org/10.1016/j.hlpt.2020.04.005 | Studied how IoT based smart disease surveillance systems can act as a potential solution to control the current pandemic | Literature Review | Limited to the theoretical model | Directing towards conducting more research on automated and effective alert systems for detection and control of the virus. |
| 2. | COVID-19: towards controlling of a pandemic **DOI**:https://doi.org/10.1016/S0140-6736(20)30673-5 | Article and WHO review for ways and recommendation on controlling COVID-19 | Literature Review | Needs of Conceptual Framework | A key outcome is to develop a framework for outbreak response. |
| 3. | The role of masks and respiratory protection against SARS-CoV-2 **DOI**: https://doi.org/10.1017/ice.2020.83 | Identifying the role of mask and personal protection again COVID-19 spread | Literature Review | Needs Empirical Evidence | N95 and Surgical mask acting are classified as key aspect to fight SARS-CoV-2 |
| 4. | Wearing face masks in the community during the COVID-19 pandemic **DOI**:https://doi.org/10.1016/S0140-6736(20)30918-1 | Studies whether only wearing mask to control spread of virus or together with social distancing and personal hygiene is also important | Literature Review | Needs Empirical Evidence | No empirical evidence of masks in infection control |
| 5. | Face Masks Against COVID-19: An Evidence Review. **DOI:**10.20944/preprints202004.0203.v1 | Study the factor of lowering community transmission using face masks | Literature Review | Needs Empirical Evidence | The review focus on providing evidence from literature to use mask for controlling spread and help frame policy around use of |

|   |   |   |   |   | non-medical masks in public |
|---|---|---|---|---|---|
| **6.** | Rational use of face masks in the COVID-19 pandemic **DOI**:https://doi.org/10.1016/S2213-2600(20)30134-X | Study the need of face mask in community settings | Literature Review | Limited to the theoretical recommendations from WHO and Health agencies. | Face masks are recommended by WHO, ICMR and other health agencies to prevent potential asymptomatic or presymptomatic transmission. |
| **7.** | Covid-19: Protecting Worker Health **DOI:**https://doi.org/10.1093/annweh/wxaa033 | Discuss use of PPE, effect of wearing mask and social distancing | Literature Review | Needs Empirical Evidence | Debates on urgent need for research on control measures to protect workers and prevent spreading |
| **8.** | Scientific and ethical basis for social-distancing interventions against COVID-19 **DOI:**https://doi.org/10.1016/S1473-3099(20)30190-0 | Discuss impact of social distancing on spread of virus | Literature Review | Needs Empirical Evidence | Focuses to create evidence-based intervention for public communication |

The past studies and research work conducted by authors hold few limitations from a conceptual framework point of view. The evidence found literature depicts the need to conduct further research focusing on devising a mature technology model infusing Artificial intelligence to respond to the outbreak in the current situation. The paper focuses on building a technology framework that can be extended for implementation in public and industrial setup to detect people with or without facemask and, therefore, eases out tracing and further control spread. Md explains that the current research work on how studying the impact of IoT based smart disease surveillance systems can help in controlling the spread of the pandemic. Siddikur Rahman,  Noah C. Peeri, Nistha Shrestha,  Rafdzah Zaki, Ubydul Haque,  Siti Hafizah Ab Hamid. (2020). Policymakers need to think about enforcing and implementing smart urban and industrial planning solutions across the geographies post-lockdown situation to resume economic activities. In the past, authors Bibri and Krogstie(2017) studies have directed towards a new generation of urban planning tools for improving mobility and accessibility; this can further be applied in combating pandemic situation across the globe.

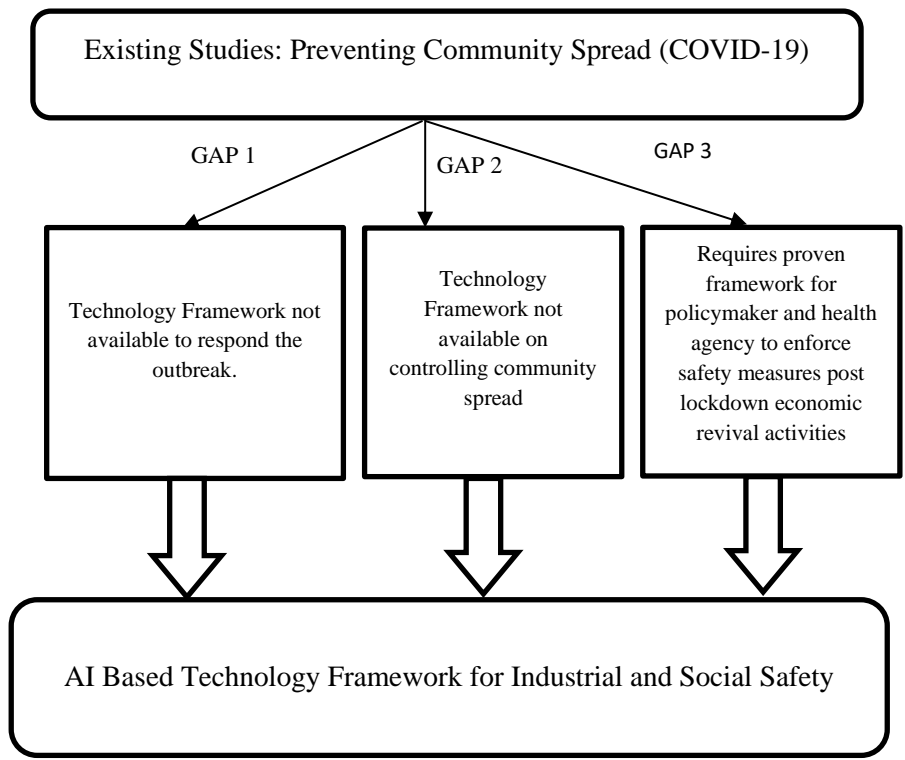

**Fig. 1.** Current Studies and GAP in outbreak response from a technology point of view for early economic revival.

## 2. The Need of technology framework

The World had seen unprecedented times before as well due to other epidemic health crises caused by Ebola, H1N1, SARS-CoV1, Zika Virus, and many others that have caused a drastic impact on the economy and social wellbeing. However, now since the exponential spread of COVID-19 globally has created a need for responses from governments due to heavy loss in GDP ranging from 3-6%, or even more depending on the country (Fernandes, Nuno, 2020) and economy predicted by the model developed by Wang et al. (2020). The section discusses the impact on the economy due to lockdown and subsequently discussion the need for a responsive framework from industrial, Citizen, and Government's viewpoint aiding to early lockdown policy formulation and strategy by Government and industries.

### 2.1 Industrial point of view

Globally, Government has taken measures to cut the spread by various measures like prohibiting gatherings of the crowd, remote health advice, creating mobile surveillance app for tracking, healthcare system to detect covid using quantum machine learning methods(Sengupta, K., Srivastava, P.R, 2021) and tracing people with the virus from studies conducted by (Keeling, Hollingsworth and Read, 2020). Though the measures that have been taken till date can help settle the pandemic stronger strategies are envisioned in reducing transmission from community and household with better support for home diagnosis facility, and dealing with the economic consequences of absence from jobs and work for individuals with experiences from past recessions researchers has suggested the impact on the economic backbone can go lower or persistent(Correia, S, S Luck, and E Verner, 2020).

Further, the discussion and studies done by (Wittkowski, 2020; Kissler et al., 2020), states that effectiveness on lockdown strategy is not known to impact the spread of the virus. Industries such as hospitality, the airline have taken a significant hit, followed by agriculture taking a global drop of more than 20% in demand, and manufacturing had shown a large drop in overall demand (Nicola et al., 2020). In the present situation with lockdown corporates and industries have started adopting digital platforms (Gaines-Ross, 2015). These industries cannot operate with work from the home policy by companies, and due to COVID-19 business has seen disruption in staffing shortage due to self-isolation and lockdown across the globe. Discussions around lifting lockdown in a phased manner and through early planning can help revive economic activities to a great extent, which later would demand effective contact tracing mechanism within industrial setup as well as community, religious places(Keeling, Hollingsworth and Read, 2020). Hence, the digital solution would play a vital role to support post lockdown phases in the overall situation, where systems like monitoring, surveillance, detection would need to be developed leveraging IoT, big data systems and AI as the core technology to be in need (Ting, D.S.W., Carin, L., Dzau, V. et al., 2020).

**2.1 Citizen's point of view**

It is important to understand the responsibility of individuals they have to play in controlling the overall spread following the guideline from local Government and WHO recommendation of practicing social distancing, personal hygiene, and wearing the mask in public places(Statista, March 31 2020b). Despite communication from health bodies, individual's awareness around the COVID-19 threat was limited, and people did not adhere to social distance as practice despite mandatory steps outlined as controlling mechanisms by WHO. Social distancing could be an effective way to reduce mortality and spread rate in any setup due to the droplet nature of the virus that can sustain in the environment(Anderseson, Heesterbeek, Klinkenberg, Hollingsworth, 2020).

The recommendation from health agencies and published government guidelines, people would strictly need to follow social distancing and follow respiratory hygiene (Leung et al., 2020) when in public. Since vaccine development is a time-consuming process, the spread can only be curb through social reformation and with individual efforts within both societal and industrial setup.

**2.2 Government's point of view**

Some Asian countries had shown success in controlling the pandemics through testing, contact tracing, and quarantine strategy along with moderate or strong social-distancing measures. China leverages intrusive surveillance technology for controlling the virus's spread tracking monitor citizens to establish safety protocol. Similarly, many other countries are now adopting use of technology in many form to communicate and trace people during this pandemic, like Ministry of Electronics and communication, India has mandated use of mobile tracking application 'Health Bridge' for contact tracking according to official website of MeitY(mygov.in). EU Members States, backed by the Commission, have rolled out mobile app 'eHealth' for automated contact tracking which is more efficient compared to manual effort which is time consuming and expensive.

The Government needs to be responsive about to pandemic situation with the citizens, create awareness campaigns, social drills through border forces, and strengthening the disasters-humanitarian coordination. Policymakers need to create effective models for forecasting, helping in making the right decisions in a timely way, even with such uncertainty around COVID-19 containment. There post lockdown strategy is a more crucial stage for Government to enforce tough decisions around individuals responding to how best to prevent transmission through governed actions by industries and individuals returning to work(Anderseson, Heesterbeek, Klinkenberg, Hollingsworth, 2020). Additionally, looking at the economic and social aspects, measures based on isolation are not sustainable in the long run. An extended drawn economic shutdown would create negative health consequences (Gilbert, M., Dewatripont, M., Muraille, E. et al., 2020). Several studies depict the role of technology it has played during earlier epidemics, to battle the current situation policymakers and health agencies need to lay a stronger documented lockdown exit strategy keeping post-implication in view and therefore leverage a strong technology framework for contact tracing and surveillance(Ting, D.S.W., Carin, L., Dzau, V. et al., 2020).

**3. Proposed Model and Framework**

In this section, the paper proposes the need for building surround outbreak response system (Fig.5), aiding in tracking and tracing safety-related concern in industrial and societal setup. The overall need for contact tracing is important to control infectious diseases and its spread (Chen, Yang, Pei, and J. Liu, 2019). The framework would encompass video feeds from a surveillance camera and IoT edge devices placed inside the industrial setup or public places to track people's movement. The architecture proposed here is a hybrid design approach to facilitate feeds from existing cameras as well as IoT devices with edge computing environments on the cloud. Light edge devices such as Intel Movidius, Nvidia Jetson, or heavy edge devices like Nvidia Tesla or Intel FPGA and Cloud Environment for training and testing large scale object detection model.

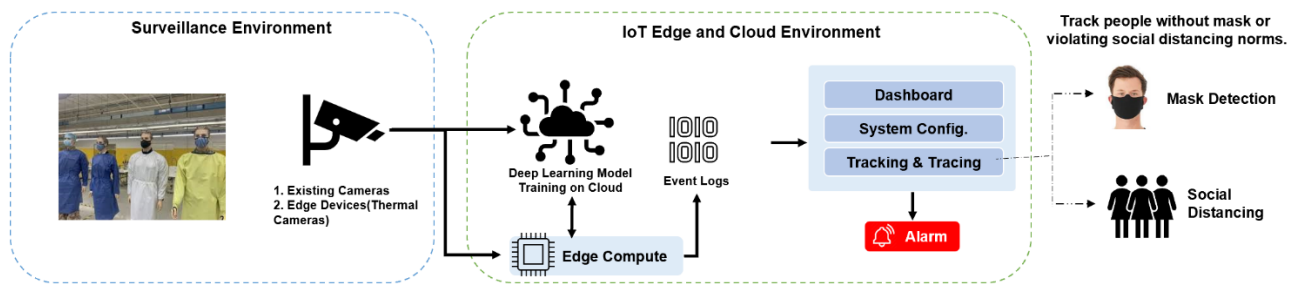

**Fig. 5.** The high-level design of outbreak response system for tracking and tracking

The framework is focused on leveraging edge computing for detecting face masks in certain environmental setup (workshops, hospitals, industrial premises). Edge computing can deliver swift localized events, near real-time insights, and reduction in overall cost due to efficient local data management and operations (Ananthanarayanan, Bahl, Cox, et al., 2019).

The positive side of deploying deep learning models on edge is to tackle bandwidth related challenges and providing extensive data security being PII in nature (Satyanarayanan et al., .2017). The edge devices can hold relatively lighter deep learning models, process raw information in a smaller size of image frames (Shi et al., 2016). The research work extends the concept of multimodal face detection and tracking of people in workshops, public and community places like temples, mosques, government offices, public transports, and offices. The paper stirs work by the researcher (Ganansia et al., 2014) using facial-recognition technique to examine the spaciotemporal behavior of individuals. Further, the research work performs advanced experiments in a controlled environment to deduce optimal algorithms for object detection and estimate social distancing criteria from image frames. The sub-sections describe the data collection strategy for the experimental design, feature engineering stages applied, and algorithm implementation benchmarking.

**3.1 System Design: Edge-Cloud Computing**

The section describes the foundation of edge computing architecture for large scale image analytics and inferencing problems. The system architecture for edge computing environment is a three-tier architecture (Ray, P.P., 2018) described in Fig. 6 and Fig 14 respectively. The key advantages of leveraging edge computing are:

- Low-latency access due to localization of compute environment, storage, and networking locally.
- Lessened bandwidth utilization due to intelligent aggregation and filtering of data to be transited for the purpose of training complex models.
- Localization of models needs intermittent access to the internet.
- Localization of Machine learning inferencing for models trained on public cloud.

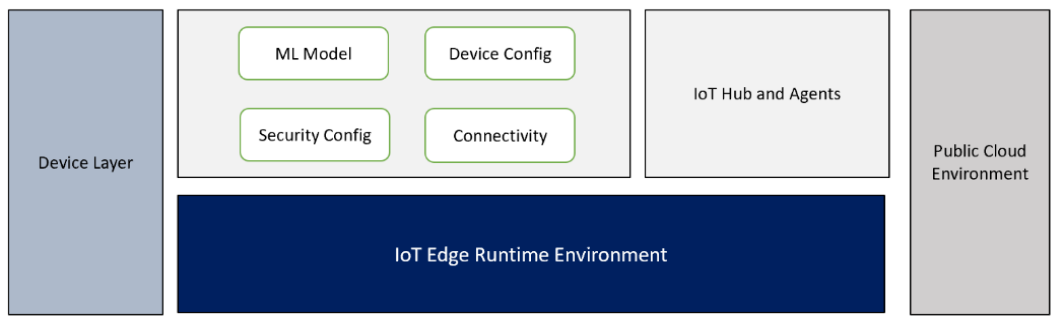

**Fig. 6.** 3-tier edge computing architecture

The architecture explains the design of Edge computing environment, the device layer connects edge devices supporting local compute to run deep learning models as compact models, provides support for connected device configuration to connect multiple mash of edge cameras, security configuration to extend the data privacy together mapped to surveillance cameras IoT edge runtime. The machine learning model training to happen in the public cloud environment and tiny trained model snapshot via docker containers would be deployed using IoT Hub and Agents within the device environment (Ananthanarayanan et al.., 2019). While Edge computing delivering resilience to overall system design from computation standpoint, the challenge of training large deep learning models and deploying for inferencing is more trivial task. The next section analyzes the relevant deep learning models for fast training and deployment with relevant benchmarking of performance and accuracy conducted on different datasets.

### 3.2 AI Models

A combinatory system using IoT, Edge, and Artificial Intelligence would provide the enterprise with a wide range of new services and business opportunities for industries and helps companies create new value(G. Hui, 2014). The working description of the model is represented as a flow diagram in Fig. 7. The model needs edge cameras deployed within the manufacturing floor shops or inside office areas continuously infer leveraging the machine learning models built in. The device capturing the area motion feeds converts video to image frames and keep in small compact patches.

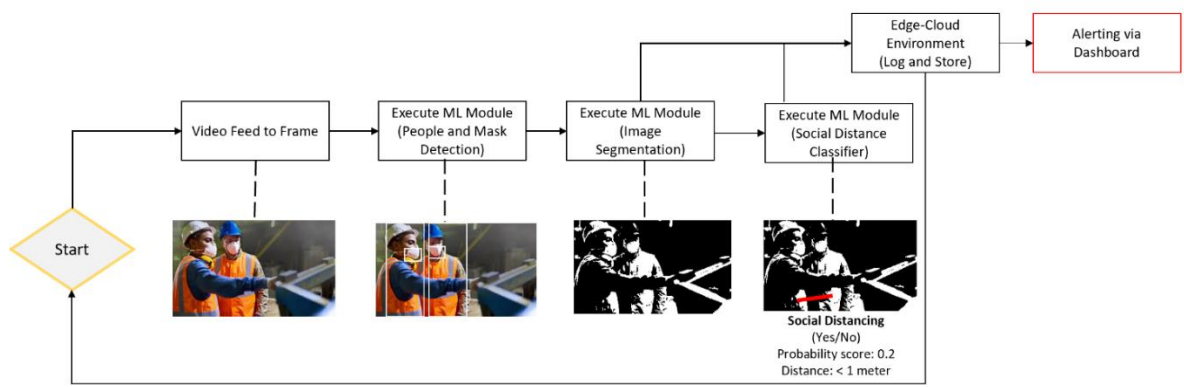

**Fig. 7.** Flow diagram of contact tracing model (Mask Detection & Social Distancing)

The role of edge cameras is to collect, and process feeds locally using the deployed DL Models. The sequence of inferencing happens at the edge environment executing people and mask detection ML module, Image segmentation ML module, and Social Distancing classifier from the segmented image. The proceed feeds are stored in a native cloud environment for further alerting and tracking mechanisms through MIS dashboards. The next subsequent sections discuss the overall system design from IT implementation perspective and algorithms evaluated for the completeness of the framework.

The application of deep convolutional neural networks has attained state-of-the-art results in solving computer vision problems like object detection, semantic Segmentation, human pose estimation (Fig. 8). The mask detection is a subset domain of object detection technique, and the social distancing is of Semantic Segmentation. The object is the problem domain to determine where objects are located within an image i.e., called object localization and which class each object belongs to called object classification(Zhao, Zheng, Xu, Wu, 2019). Semantic Segmentation dealing with the problem of assigning a class label to each pixel within an image (Chen, Wang, et al., 2019). The conceptual description of models selected for conducting the benchmarking are elucidated in subsequent sections below,

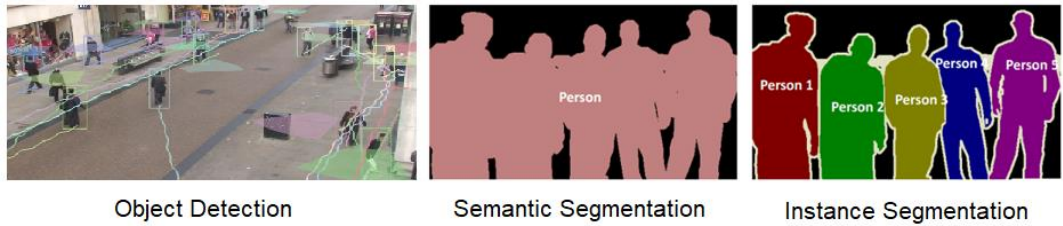

**Fig. 8.** Object Detection and Segmentation

### 3.2.1. Region based convolutional Neural Network(R-CNN)

The model uses selective search technique (Uijlings et al., 2012) in contrast of comprehensive search method in an image to detect region proposals. The initialization happens over a small region in an image merging all the regions with a hierarchical grouping. The final group are the boxes containing the entire image. The discovered regions are further combined based on color spaces and similarity metrics. The output are rare numbers of region proposals which could comprise of an object by melding small regions. How the algorithm would function is described in Fig. 9. below.

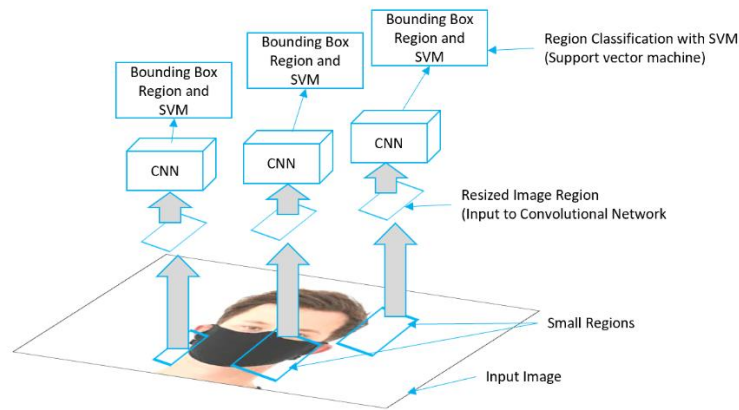

**Fig. 9.** R-CNN model

### 3.2.2. Fast R-CNN

Like R-CNN, the paper also benchmarks Fast R-CNN. The advantage of Fast R-CNN (Fig. 10.) is to reduce overall model computational expense that was happening in R-CNN version. Here, instead of multiple ConvNet being applied for each region, a single ConvNet takes the entire patch of image. A layer of regions are detected with a search method and fed into a fully connected layer creating feature vector and passed through SoftMax classifier to predict object.

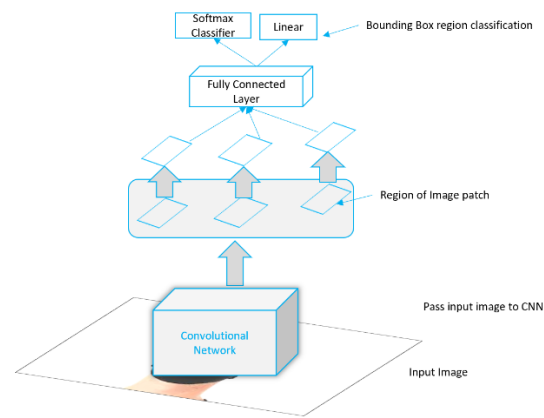

**Fig. 10.** Fast R-CNN model

### 3.2.3. Faster R-CNN

Fast R-CNN has some limitation, the search algorithm used in region detection is slow and the model replaces it with fast neural network with a novel concept of regional proposal networks. The model considers each location in previous feature-map and consider $r$ different boxes. Then from each boxes the output containing an object or not are selected and feed into Fast R-CNN.

### 3.2.4 YOLO (You only look once)

Yolo was developed by Facebook AI research group (Redmon, et al.., 2016). The architecture of Yolo is much fast that can process real-time at 45 fps. The proposed model considers object detection as a regression problem to spatially separated bounding boxes and associated class probabilities. Not Like region proposal or sliding window methods, YOLO observes an entire image during the training process and therefore implicitly encodes contextual information about classes as well as their appearance. The architecture of YOLO is described in Fig. 11. below. The method does pre-training of convolutional layers on ImageNet classification task with 224 × 224 input image resolution and thereafter fold twice the resolution for detection.

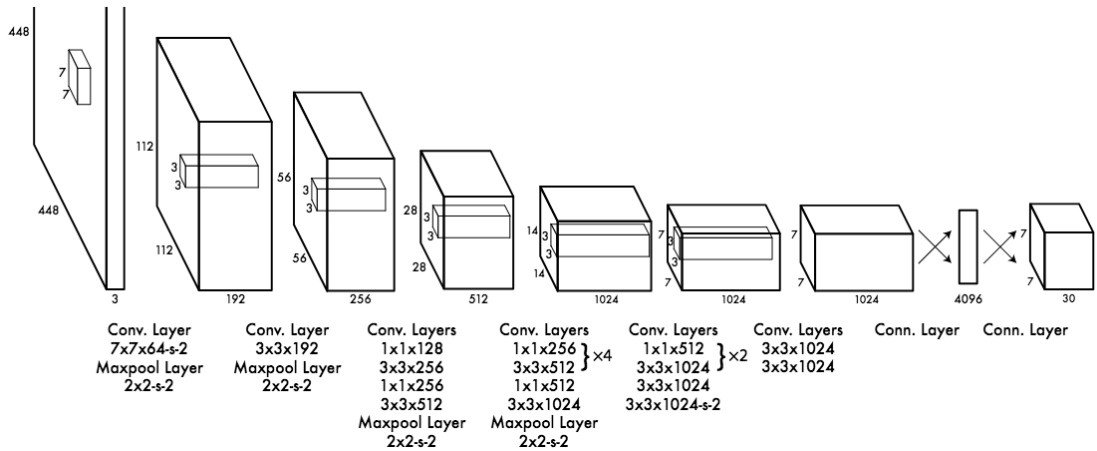

**Fig. 11.** Yolo network has 24 convolutional layers followed by 2 fully connected layers.

### 3.2.5. Semantic Segmentation using HRNetV2

The HRNetV2(Fig. 12.) recently has advanced in semantic segmentation problem, helping segment objects in an image. The method uses high-resolution representation by combining up sampled representations from all parallel convolutions in contrast to choosing representation from the high-resolution convolution in deep high-resolution representation learning model making it powerful for segmentation problems(Fig. 8). The scope of the model can further be extended to calculate distance between two segmented using distance measure functions such as Euclidian distance, hamming distance, partition distance measure functions(Porumbel, Hao, Kuntz, 2011).

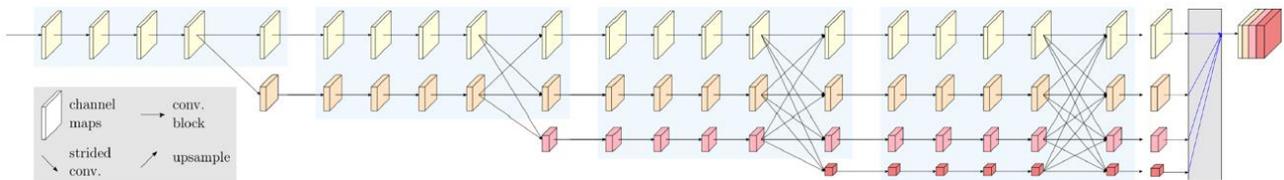

**Fig. 12.** High resolution network architecture (Sun et al.., 2019).

### 4. Methods

The section discusses the approach pursued for model development. Specifics on the overall data collection process followed by key algorithmic implementation for the study is outlined in upcoming sections.

### *4.1 Data, pre-processing and debiasing*

This section illustrates the data assimilation, pre-processing steps examined for the design of the experiments. The data collection was performed using open-source datasets repositories like msropendata.com, data.world and datasetsearch.research.google.com, pjreddie.com, town Centre dataset and image segmentation model on PASCAL VOC dataset for overall experimentations and modelling. Two sets of datasets were collected for the problem, a) images of individual personals with mask, without masks in group and unaccompanied. b) Images representing group of people in a cluster with some distance and others without distance observed. A denoising was performed on images to scale the image quality and resizing was performed for fitting the model for end experiments and create fairer model.

A known problem with image datasets is determining existence of visual representation bias of women over men. This is widespread in the history of journalism and advertising (Becker 1974; Ferree and Hall 1990; Goffman 1967). Hence, as key step during pre-processing stage, an appropriate step was taken for producing fair samples towards specific gender shade. An automated labeling tool was used to provide label to the images tagged as 'with-mask' or 'without-mask' and correspondingly label as 'following-distance' or 'not-following-distance'. The selected images from the large, collected sample corpora were first labeled using the outline tool. Post labelling a bootstrapping resampling technique was applied on the entire image dataset to set free from any representational bias(Wanyan T, Zhang J, Ding Y. Azad A, Wang Z, Glicksberg B, 2021). For gender specific biases, a debiasing method was employed to identify the male-female representation by a separate model used for identifying male and female in each image and later providing statistical estimation of frequency of the entire dataset. The process was an important step recently researchers have demonstrated how many algorithms have seen demonstrating biases(Xi, Nan et all, 2020). Authors(Crawford and Paglen, 2019) deliberately argues how ImageNet dataset that is extensively used for training image-labeling algorithms exhibit biases in more than 14+ million images collected through web scrapping from the internet.

### *4.2*  **Experiments**

The section illustrates details around the actual experimental setup and specifics related to model implementation, algorithmic implementation of social distance calculation process, and their evaluation criteria have been discusses. The experiment design phase involved selecting a base HRNET model for initial trials and rapidly performed A/B testing across other models using the similar set of datasets collected.

### *4.2.1* **Mask Detection**

The experiment adopted Yolo, Microsoft Computer Vision, Fast-RCNN, RCNN to benchmarked across samples of data chosen for identifying objects in an image. The machine learning models were trained on the collected dataset, Pascal VOC dataset and COCO dataset (Redmon & Farhadi, 2021). In previous work authors have proposed transfer learning approach(S. J. Pan and Q. Yang, 2010) fine-tuned on the MobileNetV2 model(S Yadav, 2020), however, the architecture is not tested on large scale data problems on edge systems. The paper further evaluates models for runtime as well as fastest inferencing on commercial edge devices. Here as part of experiments, Microsoft Vision AI DevKit(Figure. 13) was used to simulate the environment. The simulation environment architecture is depicted in Fig 14 below.

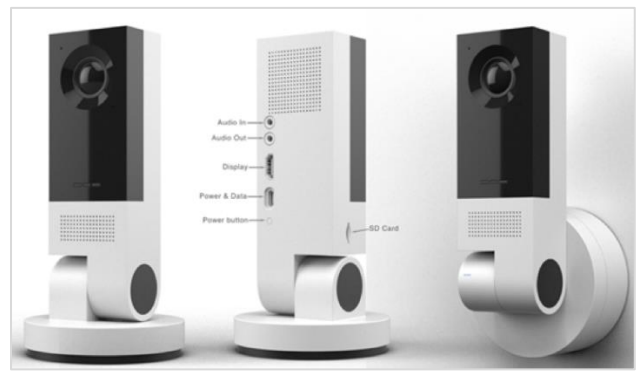

**Fig 13.** Azure IoT Vision AI Development KIT[1]

---

[1] Vision AI Development Kit - Qualcomm Developer Network

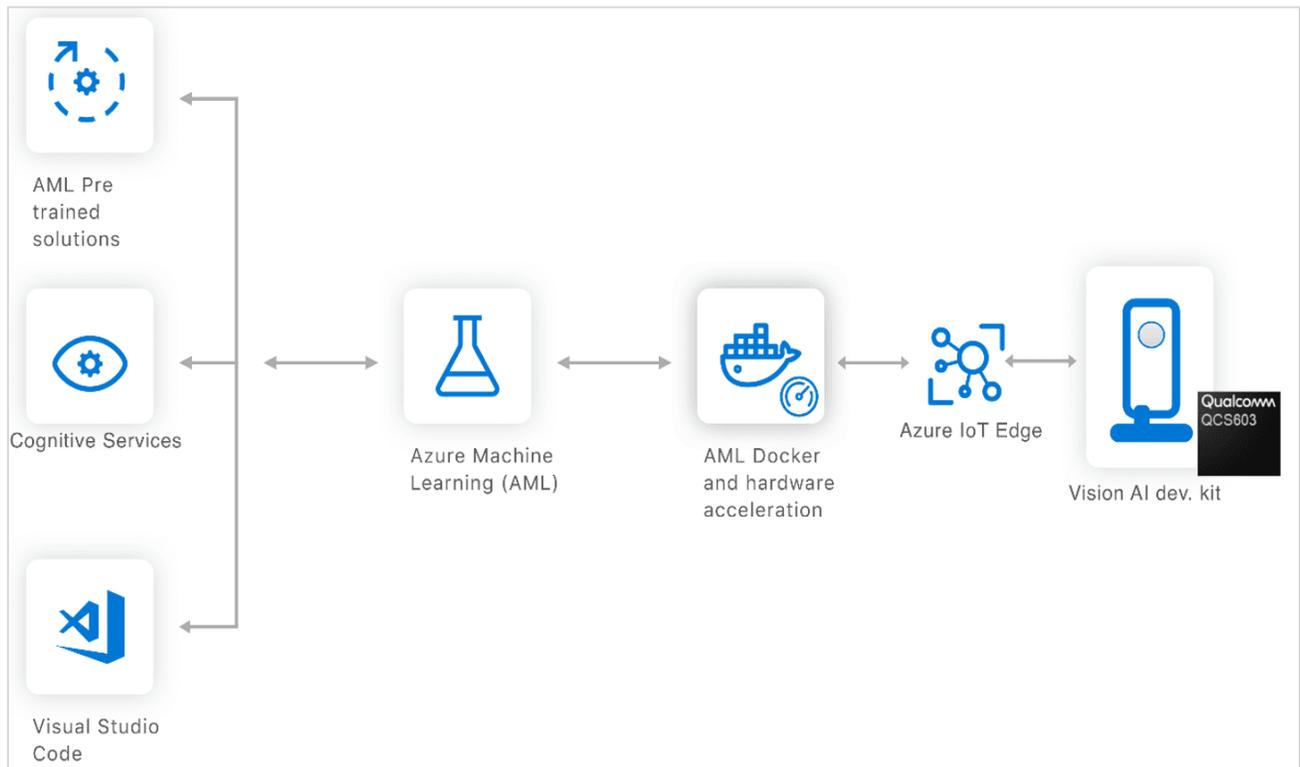

**Fig 14.** Edge Simulation architecture

The simulation tasks leveraged Vision AI dev kit to deploy custom YOLO model trained on collected datasets consisting of images of individuals with and without mask using Azure ML services. The model was further tested on real subjects in a lab environment setup for evaluation.

### *4.2.2* **Social distance calculation**

The paper introduces a distance calculation algorithm to calculate the social distance score of a segmented image using HR-Net segmentation model. The core segmentation model is developed using HRNET and Object contextual representation transformer architecture(Yuan et al., 2021). The transformer pipeline is illustrated in Fig 15. The calculation algorithm design is shown below in Table 2.

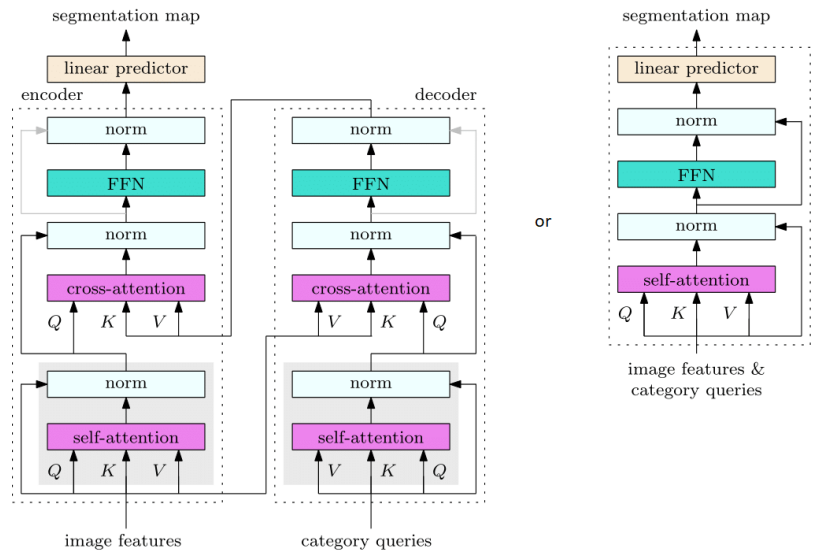

**Fig 15.** HRNET + Object Contextual representation transformer pipeline

**Table 2:** Social Distance Calculation Algorithm

| |
|---|
| **Step 1:** The model at first is set by the weights pretrained on ImageNet dataset. |
| **Step 2:** The semantic segmentation of an image frame is obtained from the above step |
| **Step 3:** The segmented images(Fig 16) is further taken as input for edge detection using Canny, dilation and erosion for removing any gap between object edge |
| **Step 4:** Detect Contours for shapes of the objects in the edge-map using findContours method in OpenCV. |
| **Step 5:** Loop over contours individually, then rotated bounding box is calculate of the contour using minAreaRect and BoxPoints method in OpenCV. |
| **Step 6:** Re-ordering the contours to organize in defined top-left, top-right, bottom-right and bottom-left order to draw the rotated bounding box and then calculate the center of the bounding box. |
| **Step 7:** To calculate distance between each object, the algorithm starts considering each contour starting with left-most as initial reference, then keeps on calculating the mid-point between top-left and top-right points followed by top-right and bottom-right points. |
| **Step 8:** In final stage, Euclidian distance is calculated between mid-points for final handling of reference object reconstruction. |

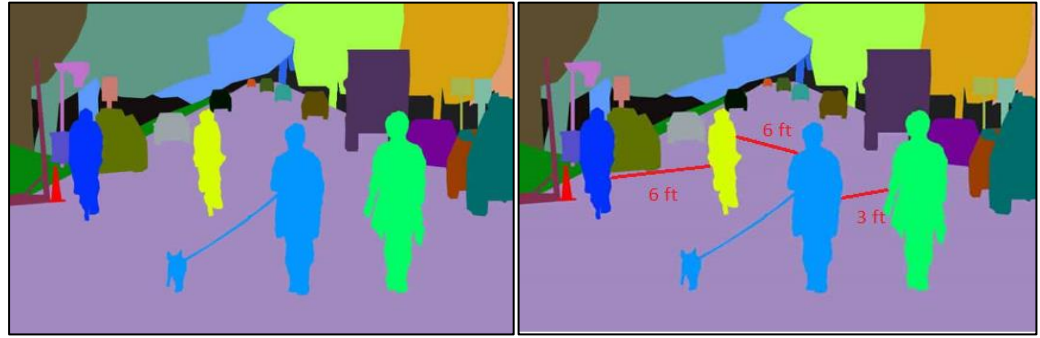

**Fig 16**. Segmented Image(left) + distance calculation between each object(right)

The designed calculation algorithms were further deployed as custom model following mask detection method in the AI dev kit environment for simulations. The scope of the simulation was limited to limited objects and human identifications in a certain environmental setup chosen for this study. The models were evaluated as cross reference benchmarking activities discussed in upcoming results sections.

## 5. Results and Discussion

The paper discusses the use of deep learning algorithms to accomplish two problems, first is detecting people with or without a mask and then predicts if social distancing is observed as image segmentation problem. The experimental results validate the performance of HRNetV2 for generating predictions for measuring social distancing score. The paper compares the computational performance of various algorithms suitable for edge inferencing and the overall accuracy of algorithms for the problem domain. The methodology adopted for benchmarking in the paper is based on secondary data collection and experimental run using existing benchmarking tools. There has been significant research work done in the space of object detection, and semantic Segmentation withing deep learning space and models like HRNET, Yolo, R-CNN, F-CNN, Mask-RCNN(refer to Table 3. for nomenclature) have proven to have yielded significant breakthrough results in image and video analytics space (Chen, Wang, et al., 2019).

**Table 3.** Terminology Classification

| | |
|---|---|
| **CNN** – Convolutional Neural Network | **FPGA** – Field Programmable gate array |
| **R-CNN** – Regions with CNN Features | **YOLO** – You only look once; An object detection system trained on COCO dataset |

| **HRNet –** High Resolution Networks | **mPA** – Mean average precision |
|---|---|
| **COCO –** Common objects in Context | **FPS** – Frames per second |
| **GPU –** Graphical processing Unit | **TP/TN** – True Positive/True Negative |
| **SGD –** Stochastic gradient descent | **DL** – Deep Learning |
| **IoT –** Internet of Things | **PASCAL-** Pattern analysis statistical modelling and computational learning |
| **FPS-** Frames per Second | **VOC –** Visual Object Classes |

Edge devices work on local inferencing or performing real-time inference on the camera itself. The technology has no transmission delay, and errors can be debug faster than the previous method, it becomes essential to evaluate DL models that are computationally inexpensive in training and deployment as well as provide better efficacy to the problem domain. The paper discusses the benchmarks for MM Detection toolbox presented in Table 4. The MM Detection toolbox is developed by Multimedia lab from The Chinese University of Hong Kong. It compares Mask RCNN and RetinaNet to help evaluate the best deep learning method to adopt for faster computing on GPU based cloud environment and edge-based local inferencing.

**Table 4.** MM Detection Analysis (Chen, Wang et al., 2019).

| **Model** | **Train (iter/s)** | **Inf (fps)** | **Mem (GB)** | **$AP_{box}$** | **$AP_{mask}$** |
|---|---|---|---|---|---|
| Mask RCNN | 0.43 | 10.8 | 3.8 | 37.4 | 34.3 |
| Mask RCNN | 0.436 | 12.1 | 3.3 | 37.8 | 34.2 |
| Mask RCNN | 0.744 | 8.1 | 8.8 | 37.8 | 34.1 |
| Mask RCNN | 0.646 | 8.8 | 6.7 | 37.1 | 33.7 |
| RetinaNet | 0.285 | 13.1 | 3.4 | 35.8 | - |
| RetinaNet | 0.275 | 11.1 | 2.7 | 36 | - |
| RetinaNet | 0.552 | 8.3 | 6.9 | 35.4 | - |
| RetinaNet | 0.565 | 11.6 | 5.1 | 35.6 | - |

The paper further contrasts overall performance of processing frames per second (highest and lowest) of various object detection models (Fig. 17) on Town Centre Dataset and image segmentation model on PASCAL VOC dataset. The visual object classes dataset is standardized image data used for object recognition problem. The benchmarks of real time systems(Redmon et al.., 2016) on PASCAL VOC 2007 and 2012 dataset is represented below in Table 5. The analysis provides higher degree of comparison on algorithms to be selected and studied further.

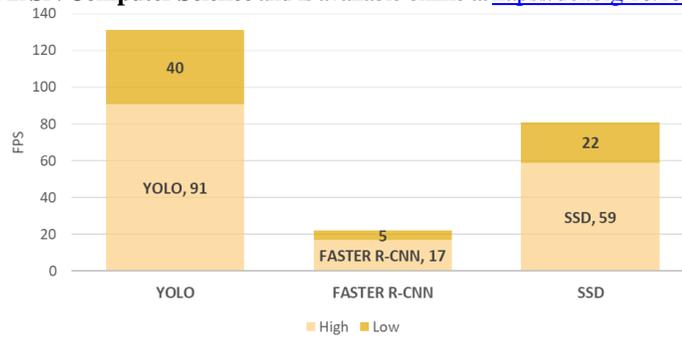

**Fig. 17.** FPS capability of various object detectors.

**Table 5.** Performance and Speed Comparison of models on PASCAL VOC dataset on Edge Compute

| Model | Speed (Real-Time) | mAP | FPS |
|---|---|---|---|
| YOLO | Yes | 63.4 % | 45 |
| Fast- YOLO | Yes | 52.7% | 155 |
| YOLO-VGG-16 | No | 66.4% | 21 |
| Fast R-CNN | No | 70.1% | 0.5 |
| Faster R-CNN VGG-16 | No | 73.2% | 7 |
| Faster R-CNN ZF | No | 62.1% | 18 |

Further to the analysis, below are training comparison conducted on Cityscape's dataset. The model training and testing were conducted with an image with input size of 512x1024 and 1024x2048. The models were set with the weights pretrained on the ImageNet on Small (Table.7) and large model (Table. 7). The analysis outcome is represented using mean Intersection over Union (mIoU). Intersection over Union evaluation metric is used for object detector on a database to measure the accuracy represented as area of overlap by area of union.

**Table 6.** Small Model

| Selected Model(s) | Number of Parameters | Multi-scale | Flip | Distillation | mIoU |
|---|---|---|---|---|---|
| ICNet | - | No | No | No | 70.6 |
| ResNet18(1.0) | 15.2 | No | No | No | 69.1 |
| ResNet18(1.0) | 15.2 | No | No | Yes | 72.7 |
| MD(Enhanced) | 14.4 | No | No | No | 67.3 |
| MD(Enhanced) | 14.4 | No | No | Yes | 71.9 |
| SQ | - | No | No | No | 59.8 |
| CRF-RNN | - | No | No | No | 62.5 |
| Dilation10 | 140.8 | No | No | No | 67.1 |
| MobileNetV2Plus | 8.3 | No | No | No | 70.1 |
| MobileNetV2Plus | 8.3 | No | No | Yes | 74.5 |
| HRNetV2-W18-Small-v1 | 1.5M | No | No | No | 70.3 |
| HRNetV2-W18-Small-v2 | 3.9M | No | No | No | 76.2 |

**Table 7.** Large Model

| Selected Model(s) | Number of Parameters | Multi-scale | Flip | mIoU |
|---|---|---|---|---|
| HRNetV2-W48 | 65.8M | No | No | 80.9 |
| HRNetV2-W48 | 65.8M | No | No | 81.2 |
| HRNetV2-W48 | 65.8M | Yes | Yes | 80.5 |
| HRNetV2-W48 | 65.8M | Yes | Yes | 81.1 |
| HRNetV2-W48 | 65.8M | Yes | Yes | 81.5 |
| HRNetV2-W48 | 65.8M | Yes | Yes | 81.9 |

The model comparison conducted encompasses small and large models. The small model is trained on smaller set of parameters ranging from 15 to 150 parameters on different models taken for benchmarking evaluated on mIoU shown in Table 6, whereas large model in Table 7, illustrates the algorithmic comparison on large parameters set 65.8 million. The analysis however limits to AI models on a limited dataset and, future empirical research work needs to be conducted to validate large scale deployment and adoption across various societal and industrial setup.

## 6. Societal implication

The situation of pandemic and Covid-19 situation across the globe has forced the government agencies, industrial ecosystem and educational setup to comply and reinforce standard safety protocols for individuals to break the spread of the virus. There has been sheer need to build technology solutions for different setup to reduce the impact of spread(Alghamdi, 2021). The study provides relevant technical solutions for policymakers and technology innovators to build systems around the problem to reduce adverse impact in post-pandemic. The study provides future directions to adopt new technology systems and help design frameworks to keep safety as key aspect while planning for return-to-work or return-to-school stages when currently both are prohibited due to surge in cases worldwide.

## 7.Conclusion

To summarize, the paper discusses an applicable framework using Artificial intelligence on edge computing for adopting an outbreak response system for contact tracing, mask detection, and detecting social distancing measures from video feeds in the surveillance systems. The benchmarks conducted on selected models to gauge the performance and accuracy yields a deeper investigational analysis to further help in the overall implementation of such a solution in industrial practice. Overall YOLO Outperforms in object detection task and is faster enough for complex edge inferencing and HRNetV2 outperform semantic segmentation problem applied to solve social distancing prediction.

The study additionally would aid in the decision making of policymakers from identified gaps in the current situation of COVID-19 spread, and how the government can formulate policy to leverage technology for contact tracking and tracing in community setup once the lockdown is lifted to resume the economy. Not only the government, but the framework would also assist and encourage industry practices to adopt and build stronger surveillance systems within the workplace environment for deeper adherence to social distancing, physical hygiene, and further thermal screening using thermal cameras to help in combating spread to a large extent.